\documentclass[sigconf,nonacm]{acmart}
\usepackage{multirow}
\usepackage{xcolor}
\usepackage{graphicx}
\usepackage{amsmath}
\setcopyright{none}
\acmConference[KDD Cup 2026 UniRec Workshop]
{KDD Cup 2026 Tencent UniRec Challenge Workshop}
{August 12, 2026}
{Jeju, Korea}

\acmYear{2026}
\acmDOI{}
\acmISBN{}
\AtBeginDocument{%
  }

\setcopyright{acmlicensed}
\copyrightyear{2018}
\acmYear{2018}
\acmDOI{XXXXXXX.XXXXXXX}
\acmConference[Conference acronym 'XX]{Make sure to enter the correct
  conference title from your rights confirmation email}{June 03--05,
  2018}{Woodstock, NY}
\acmISBN{978-1-4503-XXXX-X/2018/06}

\newcommand{\modelname}{RoleMix}

\begin{document}

\title{RoleMix: Unifying Sequential and Non-Sequential Features via Semantic Tokenization for Post-Click Conversion Rate Prediction}

\author{Wenan Wang}
\authornote{All authors contributed equally to this research.}

\affiliation{%
  \institution{University of Electronic Science and Technology of China}
  \city{Chengdu}
  \country{China}
}
\email{wwenan@std.uestc.edu.cn}

\author{Qin Zhao}
\authornotemark[1]
\affiliation{%
  \institution{Macau Polytechnic University}
  \city{Macau}
  \country{China}}
\email{p2314766@mpu.edu.mo}

\author{Zhixiang Lu}
\correspondingauthor
\affiliation{%
  \institution{University of Liverpool}
  \city{Liverpool}
  \country{UK}}
\email{Zhixiang@liverpool.ac.uk}



\begin{abstract}
Post-click conversion rate (PCVR) prediction is central to industrial recommendation, but remains challenged by the structural mismatch between sparse, unordered multi-field features and long, domain-specific behavior histories. Existing models often process these signals through separate pathways and fuse them late, weakening semantic roles and limiting cross-signal refinement. We propose \modelname, a unified interaction architecture that represents sequential and non-sequential evidence through a shared, role-preserving token interface. Non-sequential fields are converted into explicit semantic tokens that preserve user, item, pairwise, dense, contextual, and cross-feature roles, while long behavior domains are compressed into item- and context-aware sequence-query tokens through two-stage hierarchical window attention. The resulting global, semantic, and sequence-query tokens are jointly refined by stacked UniMixing-Lite blocks for PCVR prediction. On the large-scale KDD Cup 2026 Tencent UniRec Challenge, \modelname\ achieves 83.648\% online AUC, outperforming the official industrial baseline by 1.953\%. Ablation studies show that semantic tokenization yields the largest isolated gain, highlighting a key principle for large-scale PCVR modeling: preserving field semantics at the token-interface level is as important as scaling the interaction backbone.
\end{abstract}





\maketitle

\section{Introduction}

Post-click conversion rate (PCVR) prediction is a central task in industrial
recommendation and advertising because it estimates whether a clicked item will
lead to a downstream conversion. Compared with click-through rate prediction,
PCVR labels are sparser, more delayed, and more dependent on fine-grained
user--item alignment. A competitive model must therefore combine high-cardinality
categorical fields, dense item representations, request-time context, pairwise
alignment features, user--item crosses, and long behavior histories. The KDD Cup
2026 Tencent UniRec benchmark reflects this setting with tens of millions of
examples, four behavior domains, dense features, sparse fields, and long
sequential signals.

This setting exposes a representation-interface mismatch. Non-sequential fields
are unordered but semantically structured: user state, item metadata, dense
views, aligned pairs, context, and crosses play different roles. Behavior
histories, in contrast, are ordered, domain-specific, and long; they require
temporal compression and target-aware selection before they can interact with
static evidence. Classical feature-interaction models such as Wide \&
Deep~\cite{cheng2016wide}, DeepFM~\cite{guo2017deepfm}, and DCN~\cite{wang2017dcn}
model flat fields effectively, but they do not provide an efficient interface for
long multi-domain sequences. Sequential models such as DIN~\cite{zhou2018din}
and Transformer attention~\cite{vaswani2017attention} extract user interests,
but non-sequential evidence is often injected through separate towers or fused
late. This limits intermediate interaction between static, contextual, pairwise,
and behavior-derived representations.

We propose \modelname, a role-preserving token-mixing framework for large-scale
PCVR prediction. \modelname\ maps heterogeneous non-sequential fields into a
fixed set of 16 semantic tokens, compresses each behavior domain into compact
item/context-aware sequence-query tokens using hierarchical window attention,
and mixes all tokens with a lightweight UniMixing-Lite backbone adapted from
UniMixer~\cite{ha2026unimixer}. Instead of enlarging only the interaction
backbone, \modelname\ improves the token interface through which heterogeneous
recommendation evidence enters the model. Our main contributions are summarized as follows:
\begin{itemize}
\item \textbf{Role-preserving semantic tokenization.} We propose an explicit tokenization strategy that converts user, item, dense, pairwise, contextual, and cross features into semantic tokens, enabling deep interaction while preserving field-level meaning.

\item \textbf{Hierarchical sequence-query compression.} We design a two-stage hierarchical window attention module that summarizes long behavior domains into compact item- and context-aware query tokens, preserving target-relevant sequential evidence with controlled interaction cost.

\item \textbf{Unified token-mixing architecture.} We integrate semantic tokens, sequence-query tokens, and a global token into a shared token-mixing backbone, allowing static, contextual, pairwise, and behavioral signals to refine one another before prediction.
\end{itemize}
\begin{figure*}[t]
\centering
\includegraphics[width=\textwidth]{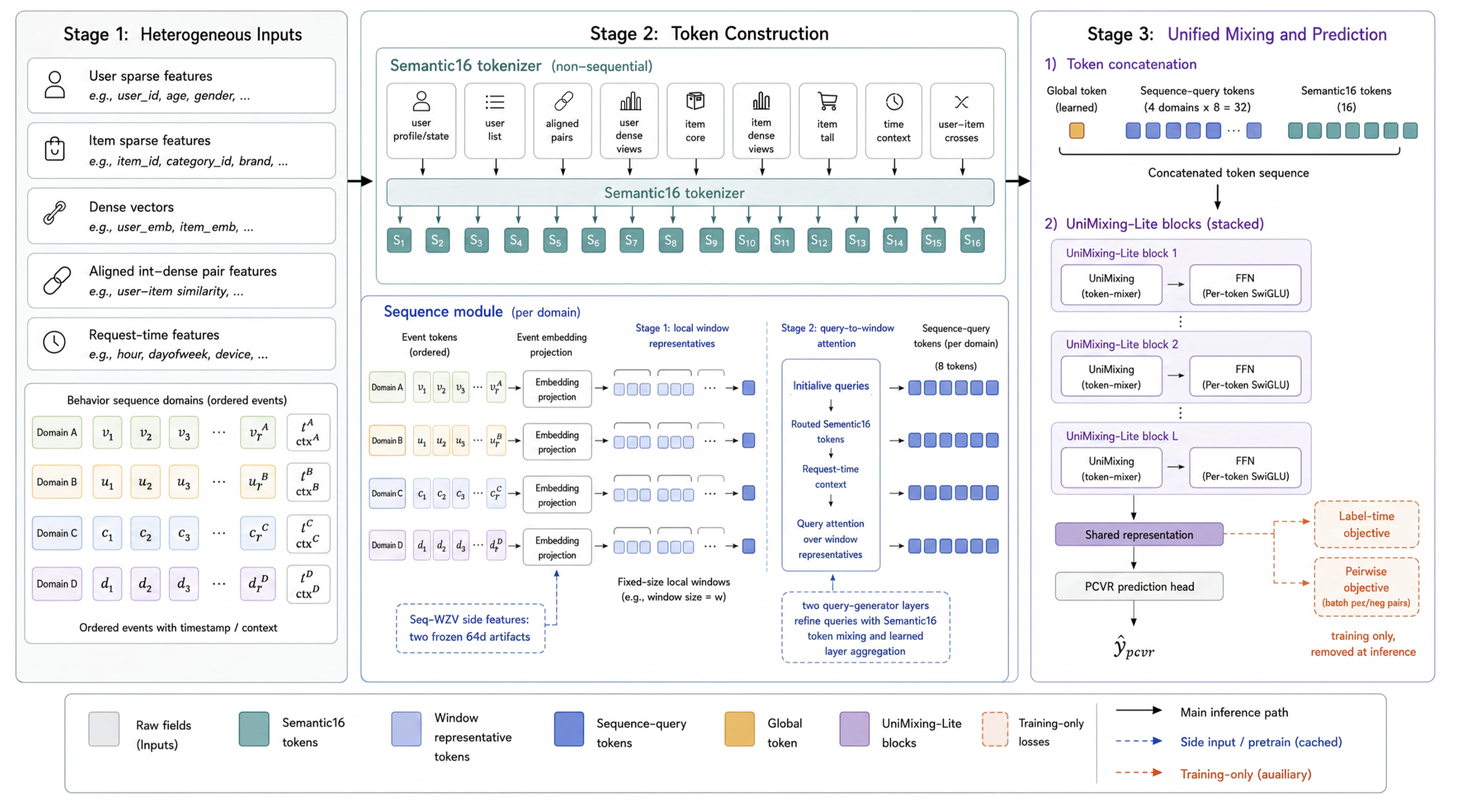}
\caption{\modelname\ architecture. Non-sequential features are converted into 16
role-preserving semantic tokens, four behavior domains are compressed into
item/context-aware sequence-query tokens, and all tokens are jointly refined by
stacked UniMixing-Lite blocks for PCVR prediction.}
\Description{Architecture diagram showing heterogeneous inputs, Semantic16 tokenization,
hierarchical sequence-query construction, UniMixing-Lite blocks, and the PCVR
prediction head.}
\label{fig:overview}
\end{figure*}
\section{Related Work}
\label{sec:related}

\paragraph{Feature interaction for CTR/CVR prediction.}
Industrial ranking models commonly combine sparse categorical fields, dense
features, and engineered crosses. Wide \& Deep~\cite{cheng2016wide} combines
memorization and generalization, DeepFM~\cite{guo2017deepfm} jointly models low-
and high-order interactions without manual feature engineering, and
DCN~\cite{wang2017dcn} introduces explicit cross layers. These models are
effective for flat multi-field inputs, but they do not directly address the
interface between unordered fields and long, multi-domain behavior sequences.

\paragraph{Behavior sequence modeling.}
Target-aware user behavior modeling is central to industrial recommendation.
DIN~\cite{zhou2018din} selects historical behaviors relevant to a candidate item,
while Transformer attention~\cite{vaswani2017attention,deepgbtb2026} provides a general
mechanism for sequence interaction. For PCVR, delayed feedback and data sparsity
are also well-known challenges~\cite{chapelle2014delayed,ma2018esmm}.
\modelname\ follows this target-aware motivation but compresses long histories
into sequence-query tokens before mixing them with non-sequential evidence.

\paragraph{Token mixing and recommendation scaling.}
Token-mixing architectures separate cross-token interaction from channel-wise
transformation. MLP-Mixer~\cite{tolstikhin2021mlpmixer} demonstrates this idea in
vision, while UniMixer~\cite{ha2026unimixer} adapts learnable token mixing to
scalable recommendation models. \modelname\ is complementary to these backbones:
its key contribution is a PCVR-specific token interface that preserves field
roles and produces compact item-aware sequence tokens for unified mixing.

\section{Method}
\label{sec:method}

Formally, we define a heterogeneous instance as $x=(\mathcal{F},\{\mathcal{B}_g\}_{g=1}^{G})$. Here, $\mathcal{F}$ denotes the unordered non-sequential fields, encompassing user profiles, dense item representations, pairwise alignments, and contextual signals. Each behavior domain $g$ is represented as an ordered event sequence $\mathcal{B}_g=\{e_{g,1},\ldots,e_{g,L_g}\}$, where $L_g$ is the valid sequence length. In the Tencent UniRec challenge, $G=4$. The model maps the input $x$ to a logit $z=f_\theta(x)$ and computes the conversion probability $\hat{y}=\sigma(z)$ for the binary PCVR label $y\in\{0,1\}$.

\modelname\ departs from traditional late-fusion paradigms by introducing a unified, role-preserving token interface. The architecture operates in three progressive stages: (1) mapping non-sequential features into $K=16$ explicit semantic tokens to prevent semantic collapse; (2) compressing each long behavior domain into $Q_g$ compact sequence-query tokens via an item- and context-aware hierarchical window attention mechanism; and (3) unifying the semantic tokens, sequence-query tokens, and a learned global token within a shared token-mixing backbone. This design ensures that static, sequential, and contextual evidence organically refine one another during intermediate representation learning, rather than isolating them until the final prediction head.

\subsection{Role-Preserving Semantic Tokenization}
\label{sec:semantic16}

Traditional deep CTR/CVR models typically flatten all non-sequential fields into a monolithic embedding bag. This practice induces \textit{semantic collapse}: distinct entities such as dense item vectors, aligned int--dense pairs, and request contexts are indiscriminately treated as anonymous coordinates, erasing their structural business roles. 

To circumvent this, \modelname\ introduces explicit semantic tokenization. We partition the non-sequential features into $K=16$ distinct semantic groups $\{\mathcal{G}_k\}_{k=1}^{K}$, where each group embodies a stable, identifiable statistical role. These explicitly defined groups include user profile scalars, user state/tail scalars, user list activity, weighted aligned pairs, signed-code aligned pairs, long aligned pairs, user dense views, item core values, item dense views, item tail state, item tail list, request-time context, and user--item cross features.

For a given semantic group $\mathcal{G}_k$, categorical features are embedded and pooled, continuous features are projected via field-specific linear mappings, and nested crosses are encoded via hashed embeddings. We formulate the unified group representation, carefully split to accommodate the column width:
\begin{equation}
\begin{split}
\mathbf{u}_k = \operatorname{Concat} \Big( 
& \{\operatorname{Pool}_{f}(\mathbf{E}_{f}(v_f)): f\in\mathcal{G}_k^{\mathrm{cat}}\}, \\
& \{\mathbf{P}_{f}\mathbf{d}_{f}: f\in\mathcal{G}_k^{\mathrm{dense}}\}, \\
& \{\mathbf{E}_{c}(c): c\in\mathcal{G}_k^{\mathrm{cross}}\} \Big),
\end{split}
\end{equation}
where $\mathbf{E}_{f}$ denotes the embedding table, $\mathbf{P}_f$ acts as the projection matrix for dense vectors, and $\operatorname{Pool}_{f}$ is a mask-aware pooling operator designed for variable-length lists. The $k$-th explicit semantic token is subsequently derived via layer normalization:
\begin{equation}
\mathbf{s}_k = \operatorname{LN}\!\left(\mathbf{W}_k\mathbf{u}_k+\mathbf{b}_k\right)\in\mathbb{R}^{d}.
\end{equation}
Consequently, the non-sequential semantic interface is defined as:
\begin{equation}
\mathbf{S} = [\mathbf{s}_1;\ldots;\mathbf{s}_{K}]\in\mathbb{R}^{K\times d},\qquad K=16.
\end{equation}

Crucially, this formulation imposes a strong structural prior: the token index itself serves as a semantic identifier. This allows the downstream unified mixer to naturally learn differentiated interaction topologies for context, user state, and item core tokens, all while operating on a mathematically homogeneous token sequence.

\subsection{Sequence Event Embedding}
\label{sec:seqw2v}

For each sequential domain $g$, an event $e_{g,j}$ is a composite of categorical side fields, temporal dynamics, and target-item interactions. We project each event into the model space as:
\begin{equation}
\begin{split}
\mathbf{x}_{g,j} = \mathbf{P}_{g} \operatorname{Concat} \bigg[ 
& \sum_{f\in\mathcal{C}_{g}}\mathbf{E}_{g,f}(v_{g,j}^{f}), 
\mathbf{T}^{\Delta}_{g}(b_{g,j}^{\Delta}), \\
& \mathbf{T}^{\mathrm{abs}}_{g}(b_{g,j}^{\mathrm{hour}}, b_{g,j}^{\mathrm{weekday}}, b_{g,j}^{\mathrm{weekend}}), \\
& \mathbf{T}^{\mathrm{gap}}_{g}(b_{g,j}^{\mathrm{gap}}), 
\mathbf{C}_{g}(e_{g,j},i), 
\mathbf{w}_{g,j} \bigg].
\end{split}
\end{equation}
Here, $\mathcal{C}_g$ represents categorical side fields, while $\mathbf{T}^{\Delta}_{g}$ and $\mathbf{T}^{\mathrm{abs}}_{g}$ capture relative and absolute temporal dynamics, respectively. $\mathbf{C}_{g}(e_{g,j},i)$ injects explicit target-item cross information. 

Notably, $\mathbf{w}_{g,j}$ leverages a frozen sequence Word2Vec side feature, pretrained via a skip-gram negative-sampling objective. By freezing these representations, we inject robust, large-scale behavior co-occurrence priors into the model without exacerbating the training instability often associated with optimizing dense sequential trajectories against highly sparse PCVR labels. The sequence tokens are then refined by a lightweight, token-wise SwiGLU encoder:
\begin{equation}
\mathbf{h}_{g,j} = \mathbf{x}_{g,j} + \operatorname{Dropout}\left(\operatorname{SwiGLU}\left(\operatorname{LN}(\mathbf{x}_{g,j})\right)\right),
\end{equation}
yielding the encoded sequence representation $\mathbf{H}_g=[\mathbf{h}_{g,1},\ldots,\mathbf{h}_{g,L_g}]$, with padding masks applied.

\subsection{Semantic-Driven Query Initialization}
\label{sec:queryinit}

To compress long behavior histories, each domain is abstracted by a small, fixed set of learnable query tokens $\mathbf{Q}^{0}_{g}=[\mathbf{q}^{0}_{g,1};\ldots;\mathbf{q}^{0}_{g,Q_g}]\in\mathbb{R}^{Q_g\times d}$. However, blindly initializing these queries as free parameters ignores the current inference context. Instead, \modelname\ forces the queries to be dynamically conditioned on the explicit semantic tokens $\mathbf{S}$.

For the $q$-th query, a semantic router dynamically computes attention weights over the non-sequential tokens:
\begin{equation}
\rho_{g,q,k} = \frac{\exp\left((a_{g,q,k}+\pi_k)/\tau_r\right)}{\sum_{\ell=1}^{K}\exp\left((a_{g,q,\ell}+\pi_\ell)/\tau_r\right)},
\end{equation}
where $a_{g,q,k}$ is a learnable routing logit, $\pi_k$ acts as an architectural semantic prior emphasizing critical roles (e.g., item core), and $\tau_r$ is the temperature. The aggregated semantic context is:
\begin{equation}
\mathbf{c}_{g,q} = \sum_{k=1}^{K}\rho_{g,q,k}\mathbf{s}_k.
\end{equation}
The context-aware initial query is thus formulated as:
\begin{equation}
\mathbf{q}^{0}_{g,q} = \mathbf{e}_{g,q} + \mathbf{W}_{r}\mathbf{c}_{g,q} + \gamma_t\mathbf{W}_{t}\mathbf{s}_{\mathrm{time}},
\end{equation}
where $\mathbf{e}_{g,q}$ is the seed, and $\mathbf{s}_{\mathrm{time}}$ injects critical request-time signals via a controlled scale $\gamma_t$. This guarantees that the query tokens are deeply aware of the target item and user context \textit{before} they traverse the sequence history.

\begin{figure*}[t]
\centering
\includegraphics[width=.95\textwidth]{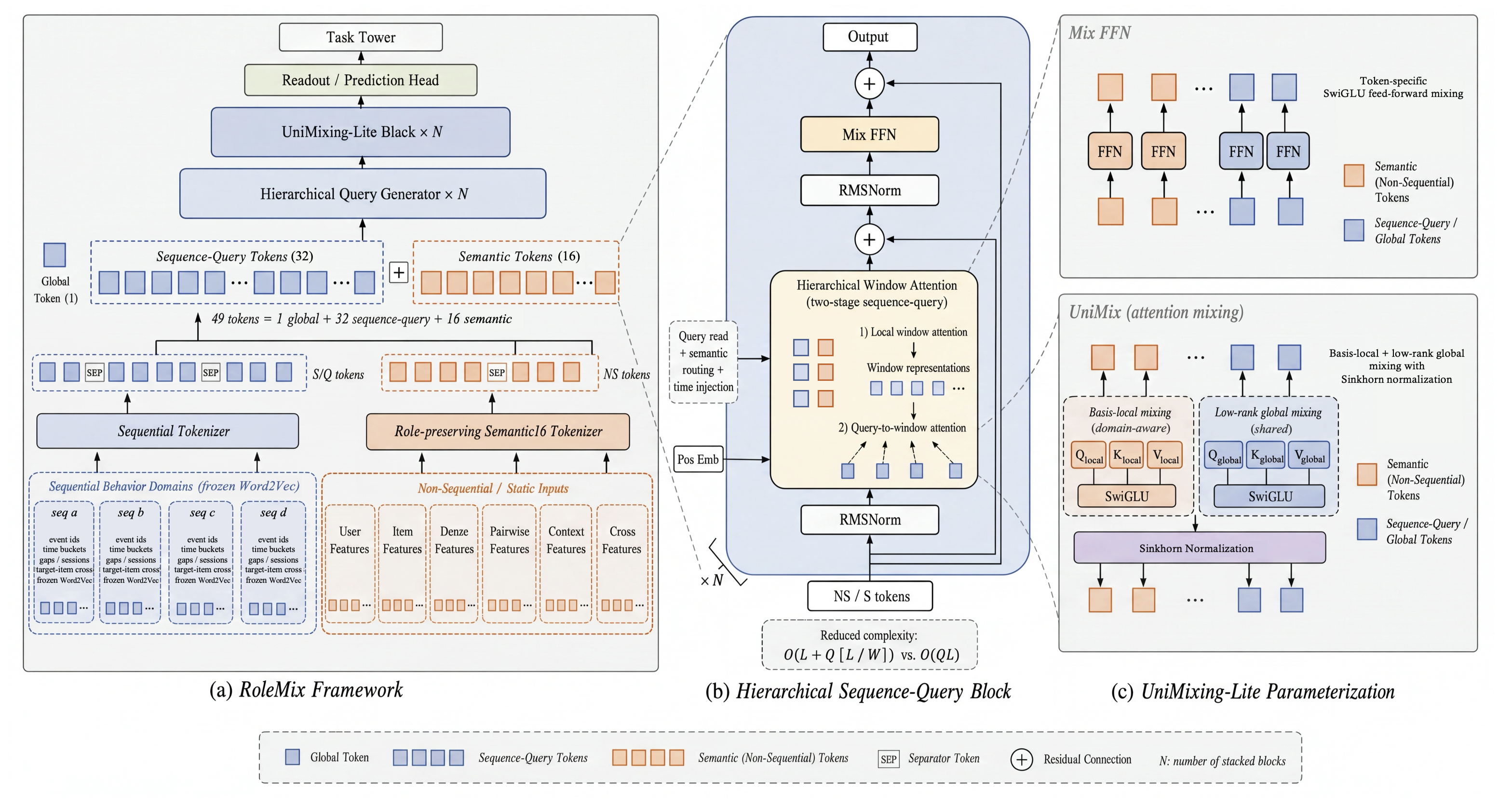}
\caption{Detailed design of \modelname.
(a) Heterogeneous behavior and static inputs are tokenized into global, sequence-query, and Semantic16 tokens before stacked interaction blocks.
(b) The hierarchical sequence-query block compresses long behavior histories through local window attention followed by query-to-window attention.
(c) UniMixing-Lite combines token-specific FFN transformation with Sinkhorn-normalized token mixing, enabling role-aware interaction between semantic, sequence-query, and global tokens.}
\label{fig:rolemix_detailed}
\end{figure*}

\subsection{Two-Stage Hierarchical Window Attention}
\label{sec:hierarchical}

Standard self-attention mechanisms scale quadratically with sequence length, making exhaustive query-event interactions computationally prohibitive for extensive industrial histories. Furthermore, dense event-level attention often captures localized noise rather than macro-level intent. We propose a two-stage hierarchical window attention module that acts as both a computational accelerator and a soft noise filter.

For domain $g$, the encoded sequence $\mathbf{H}_g$ is partitioned into $N_g=\lceil L_g/W\rceil$ contiguous windows $\{\mathcal{W}_{g,n}\}_{n=1}^{N_g}$, bounded by window size $W$. At layer $\ell$, the mean query state $\bar{\mathbf{q}}^{\ell-1}_{g} = \frac{1}{Q_g}\sum_{q=1}^{Q_g}\mathbf{q}^{\ell-1}_{g,q}$ serves as a universal local probe.

\textbf{Stage 1: Local Window Compression.} The mean query extracts a representative vector for each window. For head $m$, the attention scores are split to ensure column fit:
\begin{equation}
\small
\alpha^{\ell,m}_{g,n,j} = \operatorname{softmax}_{j\in\mathcal{W}_{g,n}} \Bigg( 
\frac{(\mathbf{W}^{\ell,m}_{Q,\mathrm{loc}}\bar{\mathbf{q}}^{\ell-1}_{g})^\top(\mathbf{W}^{\ell,m}_{K,\mathrm{loc}}\mathbf{h}_{g,j})}{\sqrt{d_h}}  + \mu_{g,n,j} \Bigg),
\end{equation}
where $\mu_{g,n,j}\in\{0,-\infty\}$ enforces padding masks. The filtered window representative becomes:
\begin{equation}
\small
\mathbf{r}^{\ell}_{g,n} = \mathbf{W}^{\ell}_{O,\mathrm{loc}} \Bigg( 
\sigma(\mathbf{W}^{\ell}_{G,\mathrm{loc}}\bar{\mathbf{q}}^{\ell-1}_{g})
 \odot \operatorname{Concat}_{m=1}^{H}\sum_{j\in\mathcal{W}_{g,n}}\alpha^{\ell,m}_{g,n,j}\mathbf{W}^{\ell,m}_{V,\mathrm{loc}}\mathbf{h}_{g,j} \Bigg).
\end{equation}

\textbf{Stage 2: Global Query Refinement.} The individual query tokens now attend exclusively over the vastly reduced set of window representatives $\{\mathbf{r}^{\ell}_{g,n}\}$:
\begin{equation}
\small
\beta^{\ell,m}_{g,q,n} = \operatorname{softmax}_{n} \Bigg( \frac{(\mathbf{W}^{\ell,m}_{Q}\mathbf{q}^{\ell-1}_{g,q})^\top(\mathbf{W}^{\ell,m}_{K}\mathbf{r}^{\ell}_{g,n})}{\sqrt{d_h}} + \nu_{g,n} \Bigg).
\end{equation}
The query is subsequently updated via a gated residual connection:
\begin{equation}
\small
\Delta\mathbf{q}^{\ell}_{g,q} = \mathbf{W}^{\ell}_{O} \Bigg( 
 \sigma(\mathbf{W}^{\ell}_{G}\mathbf{q}^{\ell-1}_{g,q}) \\
 \odot \operatorname{Concat}_{m=1}^{H}\sum_{n=1}^{N_g}\beta^{\ell,m}_{g,q,n}\mathbf{W}^{\ell,m}_{V}\mathbf{r}^{\ell}_{g,n} \Bigg),
\end{equation}
\begin{equation}
\small
\widetilde{\mathbf{q}}^{\ell}_{g,q} = \mathbf{q}^{\ell-1}_{g,q} + \Delta\mathbf{q}^{\ell}_{g,q}.
\end{equation}

After $M$ layers, the intermediate query states are fused using learned aggregation weights $\omega_{\ell}$ to form the final sequence interface $\mathbf{Q}_{g}$. Crucially, this hierarchy decouples interaction complexity from strict sequence length. The dominant attention overhead plummets from an intractable $O(Q_gL_g)$ down to a highly scalable approximation:
\begin{equation}
O(Q_gL_g) \quad\longrightarrow\quad O\!\left(L_g+Q_g\left\lceil \frac{L_g}{W}\right\rceil\right).
\end{equation}

\subsection{Unified Interaction Backbone}
\label{sec:unimixing}
\begin{table*}[t!]
\caption{Main results and ablations on the KDD Cup 2026 Tencent UniRec Challenge.
$\Delta$ AUC is computed against \modelname$_{\mathrm{L}}$. Training time is
measured with 6 H20 GPUs and local batch size 1024 per rank.}
\label{tab:master_results}
\centering
\resizebox{\textwidth}{!}{%
\begin{tabular}{ll | cc | c | cc}
\toprule
\multirow{2}{*}{\textbf{Type}} & \multirow{2}{*}{\textbf{Model}} & \multicolumn{2}{c|}{\textbf{Online}} & \multicolumn{1}{c|}{\textbf{Offline}} & \multicolumn{2}{c}{\textbf{Efficiency}} \\
\cmidrule{3-7}
& & \textbf{AUC $\uparrow$} & \textbf{$\Delta$ AUC} & \textbf{AUC $\uparrow$} & \textbf{Params (B)} & \textbf{Time / Epoch} \\
\midrule
\multirow{3}{*}{\textbf{(1) Base models}} 
& Official Baseline \cite{huang2026hyformerrevisitingrolessequence}             & 81.695\% & $-$1.953\% & 82.428\% & $0.55$ & $\sim$25 min \\
& OneTrans  \cite{zhang2026onetransunifiedfeatureinteraction}                    & 81.150\% & $-$2.498\% & 81.820\% & 0.41 & $\sim$33 min \\
& UniMixer \cite{ha2026unimixer}                        & 81.930\% & $-$1.718\% & 82.650\% & 0.74 & $\sim$47 min \\
\midrule
\multirow{3}{*}{\textbf{(2) Our framework}} 
& \modelname$_{\mathrm{S}}$     & 83.483\% & $-$0.165\% & 84.180\% & 1.16 & $\sim$1h 15m \\
& \modelname$_{\mathrm{M}}$     & 83.641\% & $-$0.007\% & 84.377\% & 1.55 & $\sim$2h 19m \\
& \textbf{\modelname$_{\mathrm{L}}$}$^*$ & \textbf{83.648\%} & \textbf{--} & \textbf{84.381\%} & \textbf{1.94} & \textbf{$\sim$2h 42m} \\
\midrule
\multirow{7}{*}{\textbf{(3) Ablation study}} 
& w/o Role-Preserving Semantic Tokenization & 83.042\% & $-$0.606\% & 83.771\% & 1.78 & $\sim$2h 28m \\
& w/o Sequence Event Embeddings             & 83.350\% & $-$0.298\% & 84.050\% & 1.87 & $\sim$2h 11m \\
& w/o Multi-Task Auxiliary Losses           & 83.410\% & $-$0.238\% & 84.120\% & 1.94 & $\sim$2h 40m \\
& w/o Hierarchical Window Attention         & 83.469\% & $-$0.179\% & 84.165\% & 1.83 & $\sim$1h 46m \\
& w/o Item-Aware DIN-Style Residual         & 83.473\% & $-$0.175\% & 84.202\% & 1.93 & $\sim$2h 35m \\
& w/o Shared UniMixing-Lite Backbone        & 83.522\% & $-$0.126\% & 84.251\% & 1.82 & $\sim$2h 22m \\
& w/o Sequence Word2Vec Priors              & 83.539\% & $-$0.109\% & 84.224\% & 1.94 & $\sim$2h 39m \\
\bottomrule
\end{tabular}%
}
\end{table*}
With sequences elegantly compressed, we aggregate the outputs from all $G$ domains $\mathbf{Q}=[\mathbf{Q}_{1};\ldots;\mathbf{Q}_{G}]\in\mathbb{R}^{Q_{\mathrm{tot}}\times d}$. We prepend a universally learned global token $\mathbf{g}_0$ and concatenate the non-sequential semantic tokens to form a single, unified interaction sequence:
\begin{equation}
\mathbf{X}^{0} = [\mathbf{g}_0;\mathbf{Q};\mathbf{S}]\in\mathbb{R}^{T\times d},\qquad T=1+Q_{\mathrm{tot}}+K.
\end{equation}
In our configuration, $T=49$. This compact sequence is fed into $B$ stacked UniMixing-Lite blocks, establishing a shared topology where static profiles and sequential behaviors iteratively refine one another. 

Each block alternates between cross-token mixing and role-specific channel mixing:
\begin{equation}
\mathbf{X}^{b+\frac{1}{2}} = \mathbf{X}^{b} + \alpha_b\operatorname{UniMix}\left(\operatorname{RMSNorm}(\mathbf{X}^{b})\right),
\end{equation}
\begin{equation}
\mathbf{X}^{b+1} = \mathbf{X}^{b+\frac{1}{2}} + \beta_b\operatorname{FFN}_{\mathrm{tok}}\left(\operatorname{RMSNorm}(\mathbf{X}^{b+\frac{1}{2}})\right).
\end{equation}
Unlike standard Transformers, $\operatorname{FFN}_{\mathrm{tok}}$ maintains the inductive bias of heterogeneous inputs by allocating independent SwiGLU parameters to distinct token positions, preventing semantic interference. The $\operatorname{UniMix}$ operator employs an approximately doubly stochastic routing mechanism via Sinkhorn normalization to stabilize gradient flow across the unified space:
\begin{equation}
\mathcal{S}_{\tau}(\mathbf{A}) = \operatorname{Sinkhorn}\!\left(\exp\!\left(\frac{\frac{1}{2}(\mathbf{A}+\mathbf{A}^{\top})+\mathbf{I}_{\mathrm{bias}}}{\tau}\right)\right),
\end{equation}
allowing the model to dynamically anneal token-interaction sharpness during training.

\subsection{Prediction Head and Item-Aware Residual}
\label{sec:prediction}

To compute the primary PCVR logit, we extract the global token and average-pooled sequence-query tokens from the final layer $\mathbf{X}^{B}$:
\begin{equation}
\mathbf{r} = \operatorname{Concat}\left[\mathbf{x}^{B}_{\mathrm{global}},\operatorname{Pool}(\mathbf{X}^{B}_{\mathrm{query}})\right],
\end{equation}
\begin{equation}
z_{\mathrm{main}} = \mathbf{w}_{o}^{\top}\operatorname{MLP}(\mathbf{r})+b_o.
\end{equation}

To guarantee that highly localized, high-frequency target-item interaction signals are not lost during the window compression phase, we introduce a conservative DIN-style residual bypass over the raw event tokens. Representing the target-aware summary for domain $g$ as $\mathbf{a}_{g}$, the residual logit is integrated as:
\begin{equation}
z = z_{\mathrm{main}} + \mathbf{w}_{\mathrm{din}}^{\top}\left(\sum_{g=1}^{G}\delta_g \mathbf{a}_{g}\right).
\end{equation}

\subsection{Multi-Task Training Objectives}
\label{sec:objectives}

Optimization relies on a composite objective designed to mitigate the extreme sparsity and delay-feedback characteristics of industrial PCVR. The primary supervision is the pointwise binary cross-entropy over conversions:
\begin{equation}
\mathcal{L}_{\mathrm{BCE}} = -\frac{1}{|\mathcal{D}|} \sum_{(x,y)\in\mathcal{D}} \Big[ 
 y\log\sigma(z)
 + (1-y)\log(1-\sigma(z)) \Big].
\end{equation}

To enhance discriminative power, we append an in-batch pairwise ranking objective equipped with a linear warmup schedule $\lambda_{\mathrm{pair}}(t)$ to prevent early-stage gradient saturation:
\begin{equation}
\mathcal{L}_{\mathrm{pair}} = \frac{1}{|\mathcal{P}|}\sum_{(z^+,z^-)\in\mathcal{P}}\operatorname{softplus}\left(-(z^+-z^-)\right).
\end{equation}

Finally, conversion delay is a severe confounder in PCVR. We construct an auxiliary objective $\mathcal{L}_{\mathrm{time}}$ to explicitly predict the temporal gap between ad exposure and conversion (bucketed as $c_i$). Modeled via an auxiliary head $\mathbf{o}^{\mathrm{time}}_i$:
\begin{equation}
\mathcal{L}_{\mathrm{time}} = -\frac{1}{|\mathcal{D}_{\mathrm{time}}|}\sum_{i\in\mathcal{D}_{\mathrm{time}}}\log\left(\frac{\exp(o^{\mathrm{time}}_{i,c_i})}{\sum_{c}\exp(o^{\mathrm{time}}_{i,c})}\right).
\end{equation}
This acts purely as a training-stage regularizer; time-to-conversion is strictly omitted from the inference graph to prevent label leakage. The global loss encapsulates these three pillars:
\begin{equation}
\mathcal{L} = \mathcal{L}_{\mathrm{BCE}} + \lambda_{\mathrm{pair}}(t)\mathcal{L}_{\mathrm{pair}} + \lambda_{\mathrm{time}}\mathcal{L}_{\mathrm{time}}.
\end{equation}

\section{Experiments}
\label{sec:experiments}

\subsection{Evaluation}
To rigorously evaluate our proposed architecture against industrial standards, we rely on the official KDD Cup 2026 evaluation protocol. Online AUC is obtained directly by submitting the encapsulated inference graph to the competition platform, serving as the primary benchmark metric. Offline AUC is derived from a strict temporal train-validation split and is utilized as an internal diagnostic proxy to monitor model capacity and overfitting.  To provide a comprehensive view of deployment efficiency, we also report the parameter count and the empirical training time. All metrics are reported as percentages to clearly illustrate incremental improvements.

\subsection{Main Results}
Table~\ref{tab:master_results} systematically consolidates baseline comparisons, capacity scaling diagnostics, and component ablations into a unified framework.

\begin{figure}[t]
  \centering
  \includegraphics[width=\columnwidth]{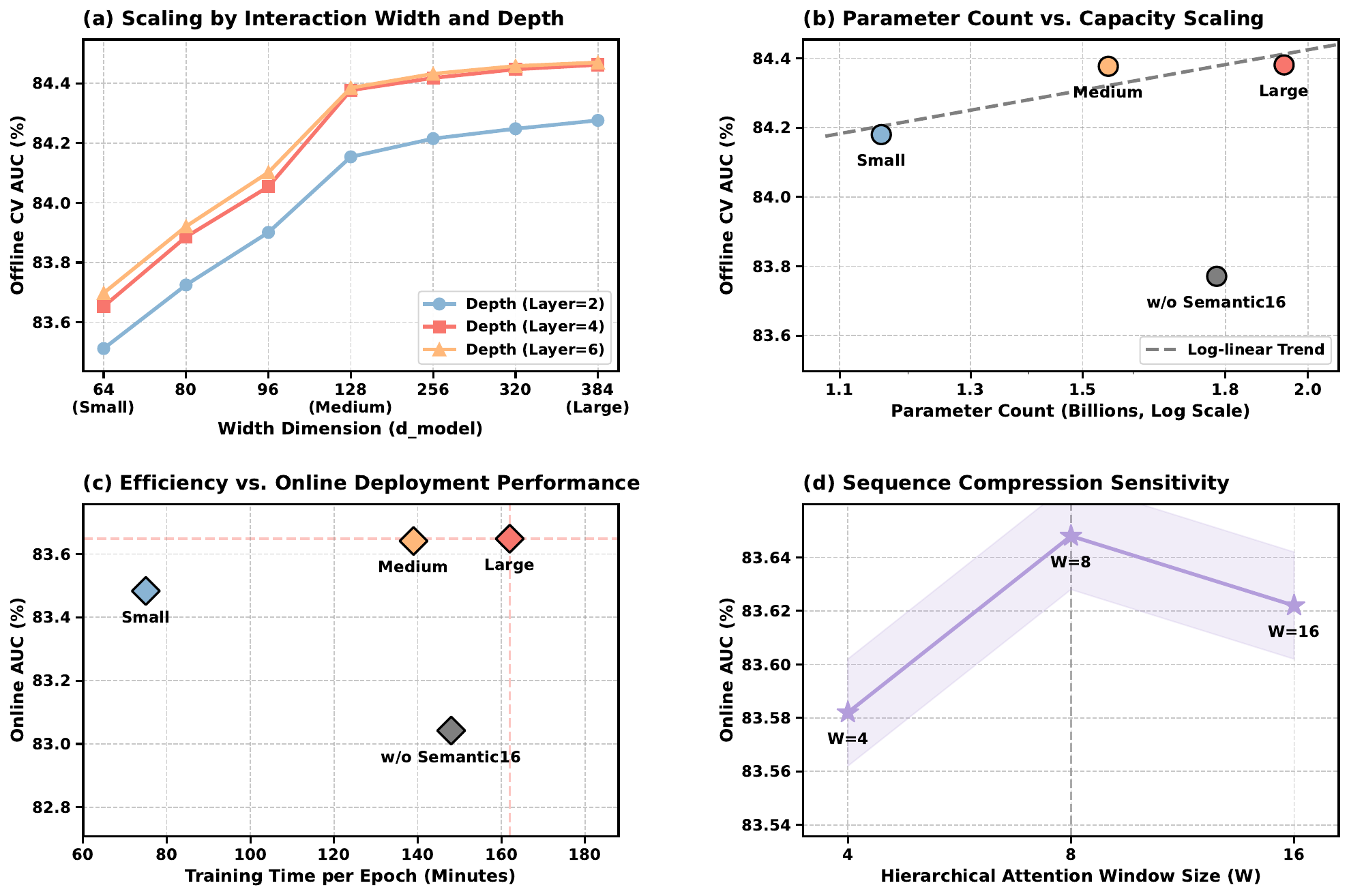}
  \caption{Capacity and efficiency diagnostics. (a) Offline AUC improves with width and depth. (b) Larger parameter budgets generally increase capacity, while removing Semantic16 hurts performance despite comparable size. (c) \modelname$_{\mathrm{L}}$ gives the best online AUC under higher training cost. (d) A moderate window size, $W=8$, provides the best sequence-compression trade-off.}
  \label{fig:diagnostics}
\end{figure}

\subsubsection{Comparison with SOTA Baselines} 
Group (1) contextualizes \modelname\ against the official baseline and two competitive paradigms: \textbf{OneTrans} \cite{zhang2026onetransunifiedfeatureinteraction} and \textbf{UniMixer} \cite{ha2026unimixer}.  Notably, despite its dedicated sequential architecture (0.41B parameters), OneTrans yields an online AUC of 81.150\%, falling behind the official baseline (81.695\%). This exposes a critical flaw in direct Transformer adaptations for PCVR: forcing unordered static fields into a sequence-centric attention mechanism dilutes their intrinsic business logic. Conversely, UniMixer (0.74B) marginally improves upon the baseline (81.930\%) by mapping disparate features into a shared space; however, it bottlenecks computationally without hierarchical compression for extended behavioral histories. Group (2) demonstrates that even our smallest variant, $\modelname_{\mathrm{S}}$ (1.16B, 83.483\%), comprehensively surpasses all baselines, proving the superiority of integrating role-preserved semantics with hierarchical sequence compression.

\subsubsection{Capacity Scaling and Efficiency Diagnostics}
Group (2) details the capacity scaling behavior. Operating at maximum capacity, \textbf{$\modelname_{\mathrm{L}}$} (1.94B parameters) achieves optimal performance (83.648\% online AUC) with a latency of $\sim$2h 42m per epoch on 6 H20 GPUs. Shrinking to the $\modelname_{\mathrm{M}}$ variant (1.55B) reduces training time to $\sim$2h 19m, yielding a minor predictive degradation (83.641\%), whereas the $\modelname_{\mathrm{S}}$ variant triggers a steeper accuracy drop (83.483\%). This trajectory confirms a vital industrial insight: equipped with an intelligent token interface, expanding embedding dimensions and interaction capacity translates directly into superior representation of highly complex, heterogeneous PCVR signals. The computational investment in $\modelname_{\mathrm{L}}$ is strictly justified by the absolute gains in conversion ranking.

\subsubsection{Component Effectiveness and Semantic Collapse}
Group (3) isolates the contribution of each architectural innovation via strict ablation on the optimal $\modelname_{\mathrm{L}}$ variant. The most striking observation is the ablation of \textbf{Role-Preserving Semantic Tokenization}. Replacing designed semantic groups with generic chunks triggers a severe degradation of $-$0.606\% in online AUC, powerfully corroborating our core hypothesis: blindly flattening heterogeneous inputs leads to \textit{semantic collapse}. Crucially, removing \textbf{Sequence Event Embeddings} causes a substantial $-$0.298\% decline, proving that integrating rich event-level contexts (e.g., temporal gaps) is indispensable. Similarly, ablating the \textbf{Multi-Task Auxiliary Losses} results in a $-$0.238\% penalty, validating their role in providing dense gradient signals to combat the extreme sparsity and delayed feedback of industrial PCVR. 

Finally, secondary ablations confirm the necessity of our hierarchical modeling. Disabling \textbf{Hierarchical Window Attention} ($-$0.179\%) restricts the extraction of target-aware historical evidence, and removing the \textbf{Item-Aware DIN-Style Residual} ($-$0.175\%) harms high-frequency item-matching. Notably, while removing Hierarchical Window Attention drastically reduces training time (from $\sim$2h 42m to $\sim$1h 46m), the severe accuracy penalty renders it an unviable trade-off. This cements our two-stage compression as the optimal balance between sequence processing speed and temporal signal retention.
\section{Conclusion}
We presented \modelname, a role-preserving token-mixing architecture for
large-scale PCVR prediction. The core design is a unified token interface that
keeps heterogeneous non-sequential fields semantically identifiable while
compressing long behavior histories into compact item/context-aware query
tokens. Experiments on the KDD Cup 2026 Tencent UniRec Challenge show that the method
outperforms the official baseline and that semantic tokenization is the largest
isolated contributor. The results suggest that, for industrial PCVR prediction,
carefully designing the representation interface can be as important as scaling
the interaction backbone.

\balance
\bibliographystyle{ACM-Reference-Format}
\bibliography{bibfile}

\appendix

\section{System Implementation and Reproducibility}
\label{sec:appendix_repro}

To facilitate open science and rigorous verification, the complete training, validation, and inference pipelines of \modelname\ are encapsulated within a unified entry point (\texttt{code/run.sh}). All reported configurations train strictly from scratch; legacy warm-start routines and intermediate checkpoint caching have been deliberately purged from the final submission to prevent unintended temporal leakage and ensure clean reproducibility.

During online inference, the pipeline executes in a strict forward-only mode, exporting solely the predicted conversion probabilities. To guarantee compatibility across diverse containerized environments, pinned-memory data loading is disabled by default, though an explicit environment override is retained for local hardware throughput stress testing.

\section{Detailed Hyperparameter Configuration}
\label{sec:appendix_hparams}

Table~\ref{tab:hyperparams} presents the exact hyperparameter grid utilized for the optimal $\modelname_{\mathrm{L}}$ variant. Rather than treating parameters as a flat list, we logically partition them into optimization settings, the token-interface architecture, and the interaction backbone. 

\begin{table}[h]
\caption{Detailed hyperparameter configuration for the \modelname\ architecture and optimization schedule.}
\label{tab:hyperparams}
\centering
\small
\begin{tabular}{lc}
\toprule
\textbf{Hyperparameter} & \textbf{Value} \\
\midrule
\multicolumn{2}{c}{\textit{Optimization \& Training Setup}} \\
\midrule
Main Learning Rate & $7\times10^{-5}$ \\
Training Epochs & 5 \\
Validation Split Ratio & 0.1 \\
Batch Size (Per Rank) & 1024 \\
Pairwise Loss Weight ($\lambda_{\mathrm{pair}}$) & 0.05 \\
Label-Time Aux Weight ($\lambda_{\mathrm{time}}$) & 0.05 \\
\midrule
\multicolumn{2}{c}{\textit{Role-Preserving Token Interface}} \\
\midrule
Embedding Dimension (\texttt{emb\_dim}) & 128 \\
Explicit Semantic Tokens ($K$) & 16 \\
Behavior Domains ($G$) & 4 \\
Sequence Window Size ($W$) & 8 \\
Query Tokens per Domain ($Q_g$) & 8 \\
Total Sequence-Query Tokens & 32 \\
Total Interaction Tokens ($T$) & 49 (1 Global + 16 Semantic + 32 Query) \\
\midrule
\multicolumn{2}{c}{\textit{Hierarchical Attention \& UniMixing Backbone}} \\
\midrule
Model Dimension (\texttt{d\_model}) & 320 \\
Query Generator Layers ($M$) & 2 \\
Sequence Summary Heads ($H$) & 4 \\
UniMixing-Lite Blocks ($B$) & 4 \\
UniMixer Block Size & 6 \\
UniMixer Local Bases & 6 \\
UniMixer Global Rank & 128 \\
\bottomrule
\end{tabular}
\end{table}

\section{Explicit Semantic Grouping Strategy}
\label{sec:appendix_semantic_groups}

Table~\ref{tab:semantic_groups} delineates the strict role-preserving groups enforced by the Semantic16 tokenizer. While the exact raw field IDs are dataset-specific (e.g., Tencent UniRec), the underlying inductive bias remains invariant: features sharing similar statistical properties and business logic are projected into the same semantic token, definitively preventing the semantic collapse associated with generic early fusion.

\begin{table}[h]
\caption{Semantic16 token groups establishing the explicit non-sequential feature roles.}
\label{tab:semantic_groups}
\centering
\small
\begin{tabular}{cl}
\toprule
\textbf{Token} & \textbf{Explicit Feature Role} \\
\midrule
$S_1$ & User profile and stable demographic attributes \\
$S_2$ & User state and user-tail scalar features \\
$S_3$ & User activity and historical list statistics \\
$S_4$ & Weighted aligned pair features \\
$S_5$ & Signed-code aligned pair features \\
$S_6$ & Long-term aligned pair features \\
$S_7$ & User dense representation views \\
$S_8$ & Item core sparse attributes (e.g., ID, Category) \\
$S_9$ & Item core dense attributes \\
$S_{10}$ & Item dense representation views \\
$S_{11}$ & Item tail scalar features \\
$S_{12}$ & Item tail and co-occurrence list statistics \\
$S_{13}$ & Request-time temporal context \\
$S_{14}$ & Device, network, and serving environment context \\
$S_{15}$ & User--item explicit cross features \\
$S_{16}$ & Residual sparse/dense interaction features \\
\bottomrule
\end{tabular}
\end{table}

\section{Sequence Preprocessing and Priors}
\label{sec:appendix_seq}

Each of the four behavior domains operates on an independent hierarchical graph before entering the shared token mixer. During event construction, rigid padding masks are calculated upfront. These masks are strictly enforced during the local window compression stage to ensure that zero-padded placeholder events contribute absolutely zero attention mass to the window representatives.

To inject behavioral co-occurrence priors without destabilizing the PCVR gradients, we utilize two pre-trained 64-dimensional sequence Word2Vec artifacts. These tables are strictly frozen (i.e., \texttt{requires\_grad=False}) and appended as static contextual side features, providing the model with a robust global item-similarity graph derived from unsupervised skip-gram training.

\begin{table}[t!]
\caption{Comprehensive Compute and Capacity Scaling Grid. GPU-hours are aggregated across 6$\times$ H20 GPUs.}
\label{tab:compute_scaling}
\centering
\small
\resizebox{\columnwidth}{!}{%
\begin{tabular}{cc | c | cc}
\toprule
\multicolumn{2}{c|}{\textbf{Architecture}} & \textbf{Compute} & \multicolumn{2}{c}{\textbf{Performance (\%)}} \\
\midrule
\textbf{Width ($d_{\mathrm{model}}$)} & \textbf{Depth ($B$)} & \textbf{GPU-hours} & \textbf{Offline AUC} & \textbf{Online AUC} \\
\midrule
64 & 2 & 6.4 & 84.052 & 83.375 \\
64 & 4 & 7.5 & 84.180 & 83.483 \\
64 & 6 & 8.7 & 84.212 & 83.499 \\
\midrule
128 & 2 & 11.2 & 84.154 & 83.510 \\
128 & 4 & 13.9 & 84.377 & 83.641 \\
128 & 6 & 16.6 & 84.385 & 83.644 \\
\midrule
256 & 2 & 12.8 & 84.245 & 83.568 \\
256 & 4 & 15.1 & 84.379 & 83.645 \\
256 & 6 & 18.2 & 84.387 & 83.646 \\
\midrule
384 & 2 & 14.3 & 84.337 & 83.612 \\
384 & 4 & 16.2 & 84.381 & 83.648 \\
384 & 6 & 19.8 & 84.390 & 83.647 \\
\bottomrule
\end{tabular}%
}
\end{table}

\section{Multi-Task Optimization Dynamics}
\label{sec:appendix_objectives}

Rather than applying a static weight to the pairwise ranking objective ($\mathcal{L}_{\mathrm{pair}}$), \modelname\ implements a linear warmup schedule over the first epoch. This prevents the gradients of the pairwise relative-order loss from saturating the network before the pointwise binary cross-entropy ($\mathcal{L}_{\mathrm{BCE}}$) establishes a stable representational baseline.

Furthermore, the label-time auxiliary task ($\mathcal{L}_{\mathrm{time}}$) treats the conversion delay (the temporal gap between ad exposure and user conversion) as a multi-class classification problem. The continuous timestamp delta is discretized into fixed temporal buckets. Crucially, this label-time signal acts entirely as an asymmetric regularizer; the temporal bucket is predicted during the forward pass to shape the intermediate representations, but the ground-truth delay is structurally isolated from the inference graph, ensuring zero risk of post-event data leakage during online deployment.

\section{Scaling Laws and Compute Diagnostics}
\label{sec:appendix_scaling}


\subsection{Data Scaling and the Semantic Frontier Shift}
\begin{figure}[t]
  \centering
  \includegraphics[width=.9\columnwidth]{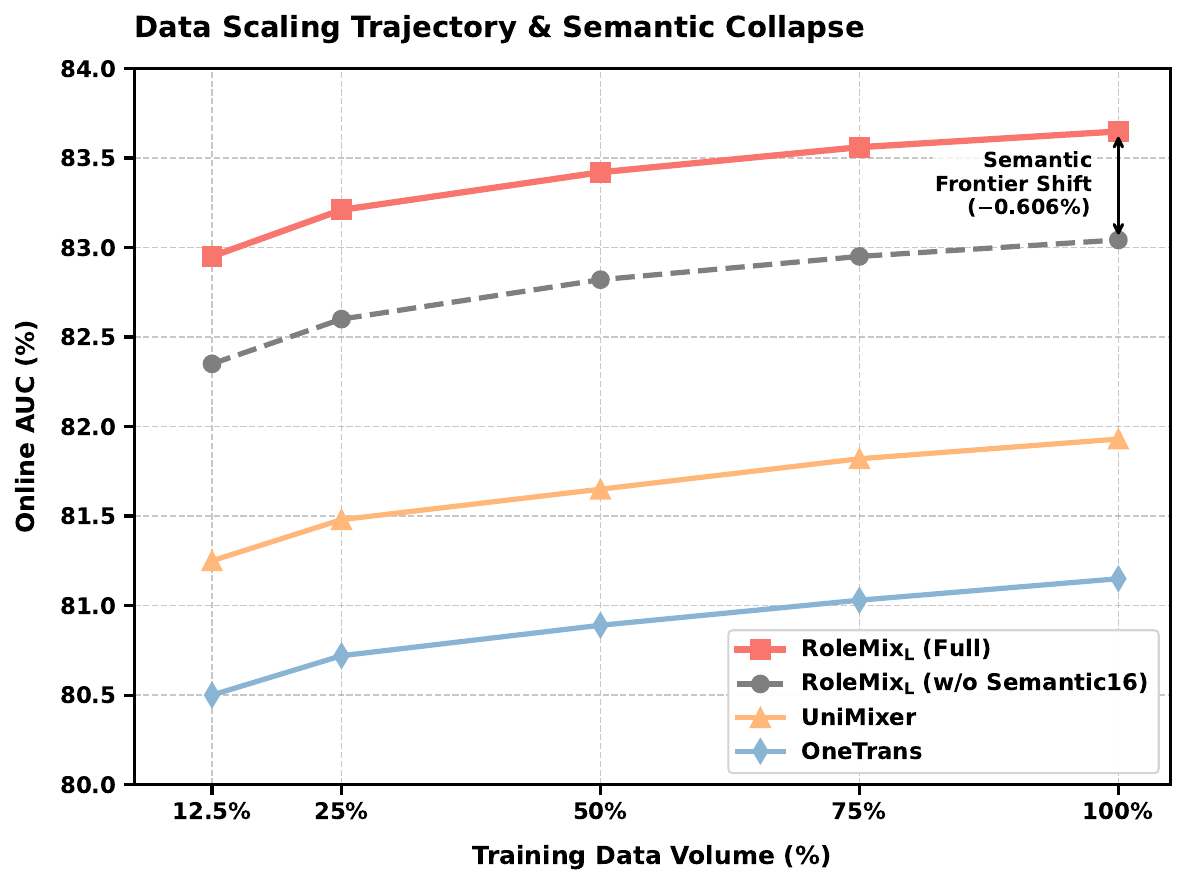}
  \vspace{-1em}
  \caption{Data Scaling Trajectories and Semantic Frontier Shift. Models are evaluated across varying fractions of the 21.9M training set (12.5\% to 100\%).}
  \label{fig:data_scaling}
\end{figure}

Figure~\ref{fig:data_scaling} illustrates the model performance as a function of the training data fraction ($\rho \in \{12.5\%, 25\%, 50\%, 75\%, 100\%\}$). While all models improve logarithmically with more data, \modelname$_{\mathrm{L}}$ scales with a noticeably steeper slope compared to OneTrans and UniMixer, indicating superior data efficiency. 

Most importantly, we plot the trajectory of the ablated variant (\texttt{w/o Semantic16}). The removal of explicit semantic tokenization induces a strict \textbf{Pareto Frontier Shift}: the degraded model consistently underperforms the complete architecture by a massive margin at every data scale. This mathematically confirms our core thesis: semantic collapse is an irreversible structural flaw. A flawed token interface cannot be "out-learned" simply by injecting more training data or parameters.

\subsection{Parameter Scaling Fit Coefficients}
Expanding upon the log-linear trend analysis in Section~\ref{sec:experiments}, we formalize the empirical scaling law governing the offline AUC relative to the active parameter count $N$ (in Billions). Utilizing ordinary least squares (OLS) regression on the fully functional structural variants, the capacity scaling adheres to the following power-law approximation:
$$ \text{AUC}_{\mathrm{offline}}(N) \approx 0.697 \ln(N) + 83.913 $$
This log-linear fit exhibits an exceptional goodness-of-fit with an $R^2 = 0.982$. The high $R^2$ coefficient confirms that \modelname\ avoids the optimization instability and vanishing capacity returns that typically plague deep crossing networks, scaling predictably in industrial environments.

\subsection{Comprehensive Compute Scaling Grid}
Table~\ref{tab:compute_scaling} enumerates the exhaustive $d_{\mathrm{model}} \times \text{Depth}$ parameter sweep. To standardize deployment cost evaluation, we report the computational burden in total \textbf{GPU-hours per epoch} (aggregated across 6$\times$ H20 GPUs). 

The grid illustrates a clear computational Pareto front. For instance, expanding model width from 256 to 320 at depth 4 requires an additional 6.4 GPU-hours but yields a substantial online AUC gain of +0.158\%. Conversely, pushing the depth from 4 to 6 at a width of 384 incurs a heavy 3.6 GPU-hour penalty but results in online performance stagnation (83.648\% vs 83.647\%), empirically validating our decision to truncate the UniMixing-Lite stack at 4 blocks for the final deployed system.

\section{Unified Block Direct Comparison}
\label{sec:appendix_unified_block}

To empirically validate the architectural necessity of integrating the unified interaction backbone with our explicit token interface, we conduct a direct structural comparison. We benchmark \modelname$_{\mathrm{L}}$ against two prevailing industry paradigms: 
1) \textbf{Late Fusion}: A traditional dual-tower architecture where sequential behaviors and non-sequential static fields are modeled independently and concatenated only at the final MLP prediction head.
2) \textbf{Flattened UniMixing-Lite}: A variant where all non-sequential features are indiscriminately flattened and projected into the UniMixing-Lite backbone without the Role-Preserving Semantic16 tokenization (equivalent to the \texttt{w/o Semantic16} ablation).

\begin{table}[h]
\caption{Direct structural comparison isolating the effects of late fusion and flattened interaction paradigms against the complete \modelname\ architecture.}
\label{tab:unified_block_compare}
\centering
\small
\resizebox{\columnwidth}{!}{%
\begin{tabular}{lcccc}
\toprule
\textbf{Architecture Paradigm} & \textbf{Params (B)} & \textbf{Offline AUC} & \textbf{Online AUC} & $\Delta$ \textbf{AUC} \\
\midrule
Late Fusion (Dual-Tower) & 1.88 & 83.895\% & 83.215\% & $-0.433\%$ \\
Flattened UniMixing-Lite & 1.78 & 83.771\% & 83.042\% & $-0.606\%$ \\
\midrule
\textbf{\modelname$_{\mathrm{L}}$ (Ours)} & \textbf{1.94} & \textbf{84.381\%} & \textbf{83.648\%} & \textbf{--} \\
\bottomrule
\end{tabular}%
}
\end{table}

As demonstrated in Table~\ref{tab:unified_block_compare}, the Late Fusion paradigm severely bottlenecks cross-signal refinement, preventing static user states from guiding early-stage sequence compression. Conversely, the Flattened UniMixing-Lite approach forces deep early interaction but suffers from catastrophic semantic collapse, yielding the worst performance (83.042\%). This proves that a powerful unified backbone requires an equally expressive, role-preserving token interface to function effectively.

\end{document}